\documentclass{article}

\usepackage{PRIMEarxiv}

\usepackage[utf8]{inputenc} 
\usepackage[T1]{fontenc}    
\usepackage{hyperref}       
\usepackage{url}            
\usepackage{booktabs}       
\usepackage{amsfonts}       
\usepackage{nicefrac}       
\usepackage{microtype}      
\usepackage{lipsum}
\usepackage{fancyhdr}       
\usepackage{graphicx}       
\graphicspath{{media/}}     
\usepackage{amsmath}
\usepackage{amssymb}
\usepackage{booktabs}
\usepackage{float}
\usepackage{caption}
\usepackage{algorithm}
\usepackage{algorithmic}
\usepackage{multirow}
\usepackage{colortbl}
\usepackage[table]{xcolor} 
\usepackage{array}
\usepackage{tabularx}
\usepackage{adjustbox}
\usepackage{siunitx}
\usepackage{cite}

\title{A Denoising Framework for Real-World Ultra-Low-Dose Lung CT Images Based on an Image Purification Strategy}

\author{
  Guoliang Gong$^{1}$, Man Yu$^{2}$, Yu Zhang$^{*}$ \\
  Tianjin University of Science and Technology \\
  \texttt{zhangyuai@tust.edu.cn} \\
   \And
  Xianghong Meng$^{3}$, Xiaoliang Wang$^{3}$, Zhongwei Zhang$^{3}$ \\
  Department of Radiology, Tianjin Hospital\\
   \\
}

\begin{document}
\maketitle

\begin{abstract}
    Computed Tomography (CT) is a vital diagnostic tool in clinical practice, yet the health risks associated with ionizing radiation cannot be overlooked. Low-dose CT (LDCT) helps mitigate radiation exposure but simultaneously leads to reduced image quality. Consequently, researchers have sought to reconstruct clear images from LDCT scans using artificial intelligence-based image enhancement techniques. However, these studies typically rely on synthetic LDCT images for algorithm training, which introduces significant domain-shift issues and limits the practical effectiveness of these algorithms in real-world scenarios. To address this challenge, we constructed a real-world paired lung dataset, referred to as Patient-uLDCT (ultra-low-dose CT), by performing multiple scans on volunteers. The radiation dose for the low-dose images in this dataset is only 2\% of the normal dose, substantially lower than the conventional 25\% low-dose and 10\% ultra-low-dose levels. Furthermore, to resolve the anatomical misalignment between normal-dose and uLDCT images caused by respiratory motion during acquisition, we propose a novel purification strategy to construct corresponding aligned image pairs. Finally, we introduce a Frequency-domain Flow Matching model (FFM) that achieves excellent image reconstruction performance. Code is available at \url{https://github.com/MonkeyDadLufy/flow-matching}. 
\end{abstract}

\section{Introduction}
\label{sec:Introduction}

COMPUTED tomography (CT) is an indispensable diagnostic tool in clinical practice, playing a vital role in disease screening, diagnosis, and treatment evaluation. During the COVID-19 pandemic, the frequency of CT examinations increased significantly, underscoring its value in crisis management.\cite{covid19-1,covid19-2} However, the ionizing radiation associated with CT scans poses non-negligible health risks, particularly for patients requiring long-term or frequent screening, such as children, pregnant women, and individuals at high risk for lung cancer.\cite{important1,important2} Studies like the National Lung Screening Trial have shown a 20\% reduction in lung cancer mortality among participants screened with LDCT.\cite{20reduced} Therefore, minimizing radiation dose per scan while preserving diagnostic integrity has become a long-standing and crucial research direction in medical imaging.

To advance LDCT denoising algorithms, the research community has developed several public datasets. Widely used datasets include MAYO2016 \cite{mayo2016}, MAYO2020 \cite{mayo2020}, the Piglet Dataset \cite{piglet}, and the Phantom Dataset \cite{phantom_dataset}. A common feature of these datasets is that their paired data are anatomically aligned. Specifically, the MAYO2016 \cite{mayo2016} dataset simulates LDCT scans by adding Poisson noise to the original normal-dose projection data (approximately 25\% for head and abdomen, 10\% for lung) and reconstructing using Filtered Back Projection (FBP). MAYO2020 \cite{mayo2020} builds upon MAYO2016 \cite{mayo2016} by providing the raw projection data; the core image data (NDCT and LDCT images) are identical. The Piglet Dataset \cite{piglet} involves scanning a deceased piglet at different dose levels (50\%, 25\%, 10\%, 5\% of normal dose). The Phantom Dataset \cite{phantom_dataset} is obtained by scanning a Gammex 467 CT phantom using a Philips Brilliance Big Bore CT scanner. These datasets, acquired under ideal conditions with perfectly aligned structures, facilitate initial algorithm validation but also contribute to performance bottlenecks when models are applied to real-world clinical data.

Despite the availability of these datasets, current research paradigms suffer from three core issues that severely limit the translation of this technology into clinical practice:

First, the research focus is significantly skewed towards achieving SOTA results on synthetic data, while exploration in the more challenging realm of uLDCT denoising remains relatively scarce. This tendency leads to numerous works pursuing marginal improvements on idealized datasets, overlooking the core needs of real-world clinical applications. For instance, on the widely used MAYO synthetic dataset, REDCNN \cite{redcnn}, as an early work, already achieved an SSIM as high as 0.97, yet subsequent research continues to seek SOTA performance on this same dataset. This over-reliance on synthetic data not only diverts attention from the clinically more valuable problem of real-world uLDCT but also introduces a second challenge.

Second, existing work interested in uLDCT primarily relies on synthetic or phantom \cite{phantom_dataset} data acquired under controlled conditions, failing to adequately address the complex challenges present in real-world clinical data. For example, while methods like DuGAN \cite{DU-GAN} and CoreDiff \cite{corediff} achieve SOTA results on uLDCT datasets like Piglet-5\% \cite{piglet}, the significant domain shift between these controlled environments and real clinical settings makes it difficult to directly apply these models in practice.

Finally, works utilizing real-world patient data typically focus on conventional LDCT and have not deeply explored the more challenging uLDCT scenario. For instance, the IPDM \cite{IPDM} method uses a private clinical dataset of normal-dose scans, but its self-supervised nature makes it difficult to generalize to the uLDCT domain with its more complex noise patterns. Conversely, the PSP\cite{psp} strategy, as a data cleaning method, tends to filter out excessive valid data when applied to extremely noisy uLDCT data, leading to insufficient model training and failure to recover critical diagnostic details.

\begin{figure*}[t]
    \centering
    \includegraphics[width=\textwidth]{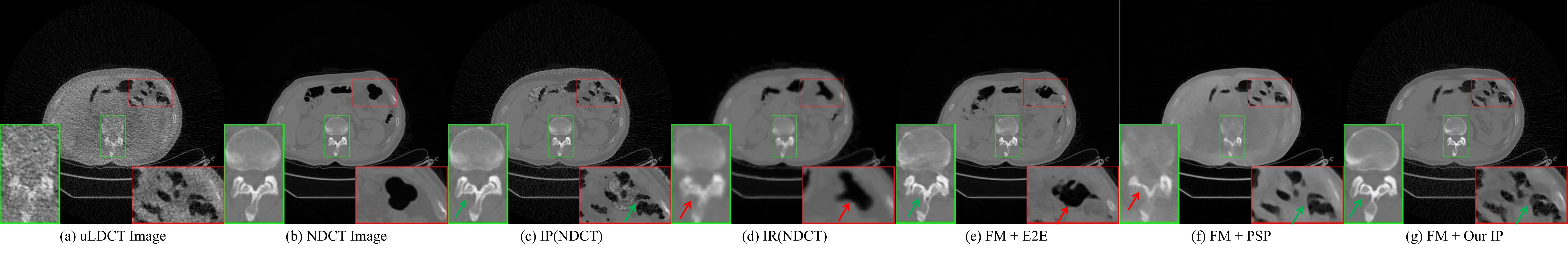}
    \captionsetup{width=\linewidth} 
    \caption{Motivation for the Image Purification (IP) strategy. (a) Ultra-low-dose CT (uLDCT) image with severe noise affecting structural clarity. (b) Corresponding normal-dose CT (NDCT) image for (a). Zoomed-in black areas show anatomical structure morphology. (c) NDCT image after IP processing, used as the label during training. (d) NDCT image after image registration. (e) Denoising result of end-to-end trained Flow Matching (FM) \cite{FM} on (a). (f) Denoising result of Flow Matching \cite{FM} trained with the PSP\cite{psp} strategy on (a). (g) Denoising result of our Flow Matching \cite{FM} trained with the IP strategy on (a). Red arrows in (c)-(g) indicate inconsistent structures; green arrows indicate consistent structures.}
    \label{fig:1}
\end{figure*}

In summary, real-world uLDCT datasets suffer from the problem of misaligned paired data. Truly exploring this domain requires first addressing this issue. We identified that slight patient motion (e.g., breathing, heartbeat) between the two scans (uLDCT and NDCT) causes severe spatial misalignment. Fig.\ref{fig:1} visually illustrates this core challenge and its impact on existing denoising methods.

Fig.\ref{fig:1}(a) shows a uLDCT image with extreme noise, severely compromising the clarity of anatomical structures. Fig.\ref{fig:1}(b) is the corresponding normal-dose CT (NDCT) image, serving as the "ground truth" reference. Comparing the zoomed-in regions (black areas) of (a) and (b) reveals clear differences in anatomical structure morphology. Directly training a network with such misaligned data pairs causes the model to learn incorrect mappings, leading to anatomical distortion in the denoised results—i.e., failing to preserve the original tissue structure of the input uLDCT image. This structural distortion can directly impact diagnostic accuracy and is a major obstacle to the clinical application of uLDCT denoising technology.

To address this spatial misalignment, researchers have explored various methods.\cite{image_registration_survey} Image registration is a classical approach in medical imaging for resolving spatial misalignment, including traditional methods like rigid \cite{Rigid_registration}, affine \cite{affine_registration}, and non-rigid deformable registration \cite{non-rigid_deformation_registration}. Recently, deep learning-based methods have surpassed traditional ones in both accuracy and speed, such as VoxelMorph \cite{VoxelMorph} and uniGradICON \cite{unigradicon}. However, for the large deformations caused by respiration and extreme noise interference between uLDCT and NDCT, these registration algorithms perform poorly and struggle to achieve precise alignment. As shown in Fig.\ref{fig:1}(d), even using the advanced uniGradICON \cite{unigradicon} for registration, the result still shows deficiencies in overall resolution (green box) and anatomical registration accuracy (red box). Recent work attempted to mitigate misalignment by filtering out misaligned image patches via data cleaning. However, on extremely noisy uLDCT data, this method filters out too much data, leading to under-training. As shown in Fig.\ref{fig:1}(f), the flow matching model \cite{FM} trained with the PSP\cite{psp} strategy produces denoised results with severe structural blurring and loss of detail.

Fig.\ref{fig:1}(e) and Fig.\ref{fig:1}(g) further compare the denoising effects under different strategies. Fig.\ref{fig:1}(e) shows the denoising result of applying an end-to-end trained flow matching model \cite{FM} to (a). The red arrows indicate regions where the structure is inconsistent with the original uLDCT, showing the model's failure to preserve the original anatomical contours. In contrast, Fig.\ref{fig:1}(g) shows the denoising result of flow matching \cite{FM} trained with our proposed IP strategy. The green arrows indicate regions where the structure is consistent with the original uLDCT, demonstrating the exceptional ability of the IP strategy to maintain anatomical integrity. Fig.\ref{fig:1}(c) shows the NDCT image after processing by the IP strategy, which serves as a structurally aligned label during training and is key to achieving high-quality denoising.

To address the aforementioned problems, this paper proposes a systematic denoising framework centered on an Image Purification strategy. This strategy operates at the image level, effectively correcting structural misalignment between uLDCT and NDCT images to generate high-quality training data. Fig.\ref{fig:3} illustrates our overall framework, which comprises three decoupled stages: data preprocessing based on the IP strategy, FFM model training and sampling, and model evaluation. The main contributions of this paper are as follows:

\begin{itemize}
    \item  We constructed a real clinical uLDCT lung dataset with a radiation dose of only 2\% of the normal level, providing a data foundation for related research.
    \item We propose an innovative Image Purification strategy that effectively resolves the spatial misalignment problem in real-world uLDCT-NDCT data pairs.
    \item We propose a Frequency-domain Flow Matching model, which, combined with the IP strategy, achieves SOTA performance on uLDCT denoising.
\end{itemize}

The remainder of this paper is organized as follows: Section \ref{sec:RELATED WORK} reviews related work. Section \ref{sec:PROPOSED METHOD} details our proposed denoising framework. Experiments in Section \ref{sec:EXPERIMENTS} demonstrate that our IP strategy quantitatively and qualitatively enhances seven LDCT denoising networks, and our Frequency-domain Flow Matching model achieves SOTA results in anatomical contour preservation. Section \ref{sec:CONCLUSION} concludes the paper.


\section{RELATED WORK}
\label{sec:RELATED WORK}

\subsection{Methods for Addressing Data Misalignment}
\label{sub:解决数据对不对齐的方法}
Image registration is a classical approach in medical imaging for resolving spatial misalignment, focusing on finding an optimal spatial transformation model for geometric alignment between images. Traditional methods, listed in increasing order of model complexity, include: Rigid registration \cite{Rigid_registration}, which permits only rotation and translation, suitable for aligning image sequences of the same body part where internal structures are assumed unchanged, e.g., skull bones; Affine registration \cite{affine_registration}, which introduces scaling and shearing atop rigid transformations, capable of compensating for overall object size changes and imaging perspective differences, often used as a preprocessing step for complex registration, applicable for initial alignment of brain images from different subjects; Non-rigid deformable registration\cite{non-rigid_deformation_registration}, which describes local, free, smooth deformations via dense displacement fields, capable of simulating organ physiological motion (e.g., respiration, heartbeat) and tissue elasticity changes. It is a key technique for registering abdominal, pulmonary, and other body parts, but is computationally complex and prone to non-physical deformations. Recently, the rise of deep learning has catalyzed a paradigm shift from "iterative optimization" to "forward prediction," yielding efficient algorithms like VoxelMorph \cite{VoxelMorph}, which uses a convolutional neural network to directly predict dense deformation fields end-to-end, achieving registration speeds orders of magnitude faster than traditional methods and enabling near-real-time applications. uniGradICON\cite{unigradicon}, pre-trained on a mixed dataset of multiple anatomical structures and modalities, and employing gradient inverse consistency constraints to eliminate dependence on task-specific hyperparameters, empowers a single model with strong zero-shot generalization capability, applicable directly to unseen datasets without fine-tuning. However, despite the excellent performance of these advanced algorithms, they still underperform when dealing with the large deformations from respiration and extreme noise interference between uLDCT and NDCT. Severe noise hinders the network's ability to extract effective features, while large deformations impose extremely high demands on the regularization of the deformation field, making it difficult to achieve registration that is both accurate and topology-preserving. As shown in Fig.\ref{fig:1}(d), even with the advanced uniGradICON \cite{unigradicon} model, the registration result exhibits noticeable deficiencies in overall resolution (green box) and anatomical structure accuracy (red box).

The work \cite{psp} proposed a data cleaning strategy called PSP (Patch Similarity Purification). It divides each 512$\times$512 image into 64 patches of 64$\times$64 pixels and filters out patches where the LDCT and NDCT patches are dissimilar; the remaining patches are used for training. However, this strategy faces significant challenges when applied to our uLDCT lung dataset. The extreme noise and poor structural visibility in uLDCT images lead to an excessively high proportion of patches judged "dissimilar," resulting in severely insufficient training data and ultimately poor denoising performance (Fig.\ref{fig:1}(f)). In conclusion, both traditional image registration methods and emerging data cleaning strategies reveal inherent limitations when addressing the data misalignment problem in the extreme scenario of uLDCT. This highlights that developing a new method capable of effectively overcoming the inherent spatial misalignment in real uLDCT-NDCT data pairs is a key challenge for the clinical translation of uLDCT denoising.

\subsection{Low-Dose CT Image Denoising Networks}
\label{sub:低剂量CT图像去噪网络} 
Existing CT denoising networks are largely tailored for conventional LDCT scenarios, and their methodologies can be categorized into three main paradigms: The first aims to separate and remove noise $n$ from the noisy image$y = x + n$ to recover the target NDCT image $x$. The second focuses on building an end-to-end mapping $f(\cdot)$ from the LDCT image $y = f(x)$ to the NDCT image $x$. The third adopts a self-supervised learning paradigm by modeling the mapping $g(\cdot)$ for $\hat{y} = g(y)$ or $\hat{x} = g(x)$, and using $g(y)$ or $g^{-1}(\hat{x})$ to approximate the NDCT image $x$.

Among the first category, the RED-CNN network achieved remarkable results with its U-Net architecture. The EDCNN \cite{edcnn} network introduced an edge enhancement module to improve the fidelity of anatomical contours in denoised images. DU-GAN utilized a U-Net-based discriminator to learn global and local differences between denoised and normal-dose images in both image and gradient domains, thereby enhancing edge information. MMCA \cite{MMCA} adopted a Vision Transformer (ViT) \cite{vit} macro-architecture combined with Sobel operators to enhance edge features. Although these models perform reasonably well on conventional LDCT denoising tasks, their ability to reconstruct fine texture details significantly diminishes in the uLDCT scenario. More critically, despite design considerations for enhancing anatomical contours, the inherent "misalignment" in the training data often prevents the denoised results from faithfully preserving the original tissue structure of the input uLDCT image, posing a potential threat to diagnostic accuracy.

\begin{figure*}[htbp]
    \centering
    \includegraphics[width=\textwidth]{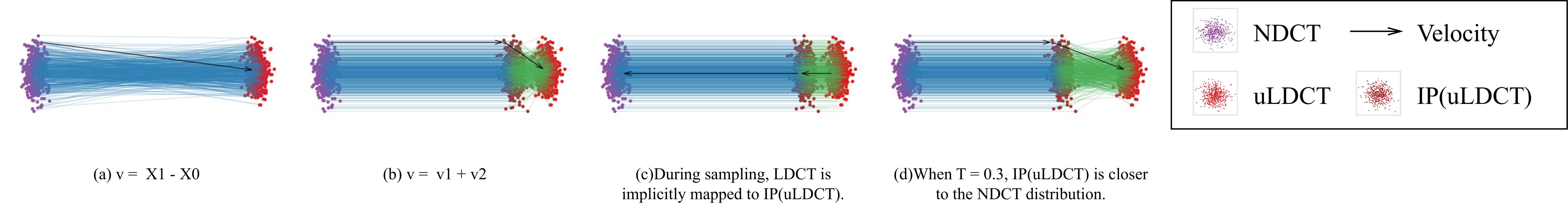}
    \captionsetup{width=\linewidth} 
    \caption{Concept of the Image Purification (IP) Strategy. \textbf{(a)}Mapping relationship of input data($X_0$,$X_1$) $\sim$ (NDCT,uLDCT).Structural misalignment between $X_0$ and $X_1$ leads to intersecting trajectories.  \textbf{(b)}By establishing a new distribution $\mathrm{X}_1{ }^{\prime} \sim IP(uLDCT)$,decompose $v$ into structure-consistent $v_1$ and texture-consistent $v_2$.Use $v_1$ as the training data for the model.  \textbf{(c)}During inference, $X_1$ is implicitly mapped from uLDCT to IP(uLDCT), and then reaches NDCT via the trained velocity field. \textbf{(d)}Parameter T influences the magnitude of $v_1$.}
    \label{fig:2}
\end{figure*}

The second category comprises mapping methods based on diffusion models, which have recently shown great potential. Cold Diffusion \cite{coldDiffusion}
generalized traditional diffusion models by modifying the noise addition mechanism to handle arbitrary image transformations. Building on this, CoreDiff \cite{corediff} further improved performance by introducing an error regulation recovery network to correct sampling errors. Flow Matching \cite{FM} employed Optimal Transport \cite{Optimal_Transport} displacement interpolation to define conditional probability paths and utilized numerical ODE solvers for fast sample generation. However, the Rectified Flow \cite{reflow} points out that the mapping paths constructed by Flow Matching \cite{FM} contain numerous crossing points (Fig.\ref{fig:2}(a)). These crossings significantly increase the difficulty of network sampling. Although techniques like model distillation can straighten the paths, this merely establishes a mapping between uLDCT and NDCT without ensuring that the mapping is "structure-preserving." During training, the model is not explicitly constrained to preserve the anatomical structure of the input uLDCT image, easily leading to tissue structure distortion in the output. This paper further argues that the misalignment between uLDCT and NDCT images is also a significant cause of crossing in the mapping paths, thereby affecting the model's sampling performance. Resshift \cite{resshift} applied Optimal Transport \cite{Optimal_Transport}  displacement extrapolation to latent diffusion models, significantly reducing the number of sampling steps and paving the way for practical deployment. Although these diffusion-based networks can effectively restore texture details in uLDCT, the anatomical distortion problem caused by "misaligned" training data persists and remains fundamentally unsolved.

The third category, self-supervised methods, has garnered significant attention due to their independence from paired data. The N2N \cite{N2N} method uses only LDCT data, employing a self-supervised noise2noise strategy. It adds noise multiple times to the same LDCT image and uses pairs of singly and doubly corrupted images instead of NDCT-LDCT pairs for training. The IPDM \cite{IPDM} method uses only NDCT data, adding noise to NDCT images up to an early stage of the diffusion process to generate LDCT-like images for training. However, these methods typically assume that the noise in LDCT follows a known statistical distribution, which deviates from the reality in real-world uLDCT images where noise is highly coupled with anatomical structure. Consequently, self-supervised methods struggle to correctly remove the complex noise present in real uLDCT images.

\section{PROPOSED METHOD}
\label{sec:PROPOSED METHOD}

\begin{figure*}[htbp]
    \centering
    \includegraphics[width=\textwidth]{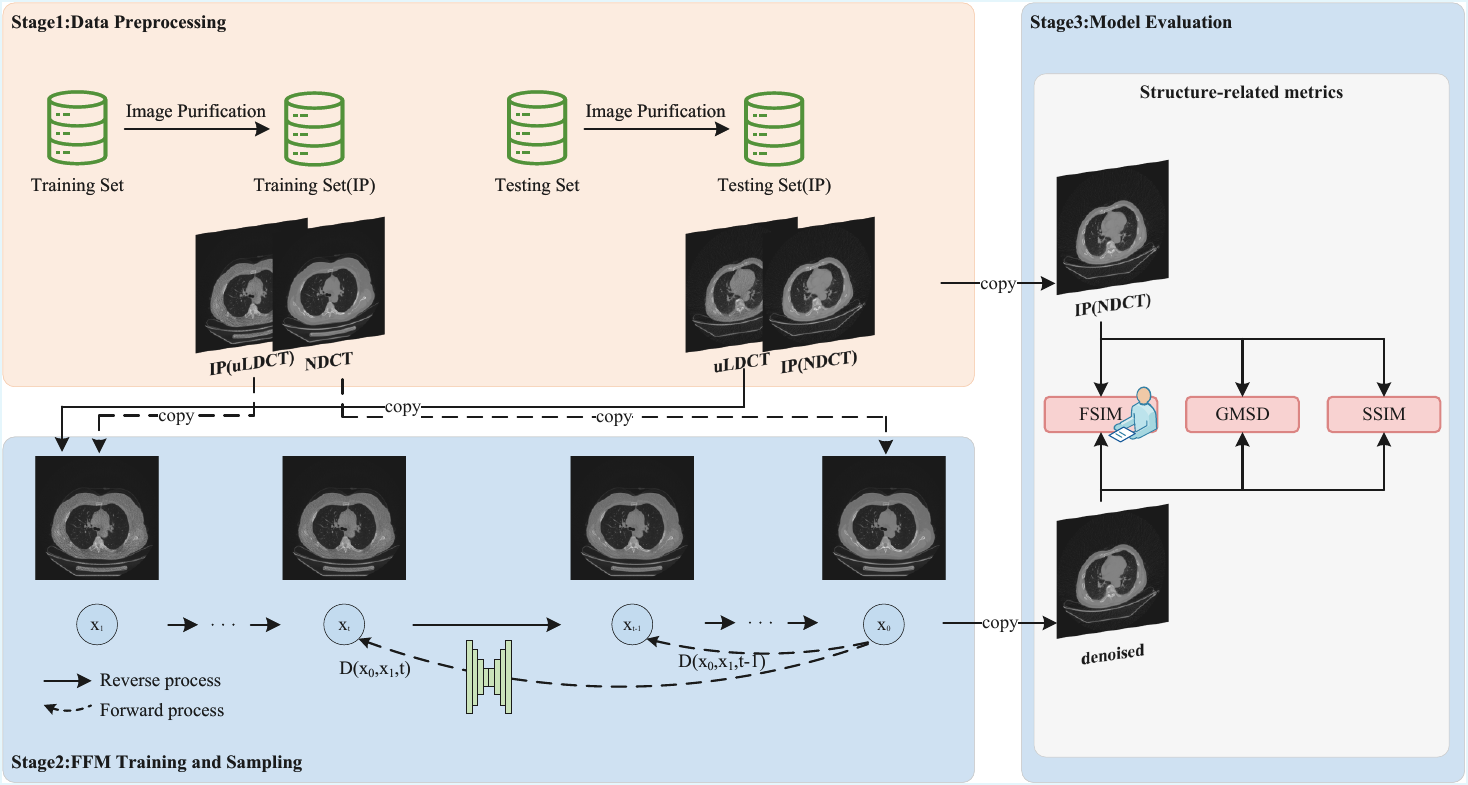}
    \captionsetup{width=\linewidth} 
    \caption{Overall Framework.}
    \label{fig:3}
\end{figure*}

To address the aforementioned challenges, we propose a systematic denoising framework centered on an innovative Image Purification strategy. As shown in Fig. \ref{fig:3}, the framework comprises three decoupled stages: Data Preprocessing, Model Training, and Evaluation. This modular design ensures the independence and replaceability of each component.

\subsection{Image Purification Strategy}
\label{sub:图像纯化策略}
Our Image Purification (IP) strategy aims to resolve the spatial misalignment between uLDCT and NDCT images. Its core concept is illustrated in Fig.\ref{fig:2}. From a probabilistic modeling perspective, data misalignment leads to numerous crossings in the mapping paths from the uLDCT distribution to the NDCT distribution (Fig.\ref{fig:2}(a)), making the learning process of the denoising model difficult and unstable. The core of the IP strategy is to construct an intermediate distribution—IP(uLDCT)—during the training phase that is structurally aligned with the NDCT distribution. This creates a more direct and easier-to-learn mapping relationship on the probability path (Fig.\ref{fig:2}(b)). During inference, the input uLDCT is automatically mapped to the IP(uLDCT) distribution and then proceeds to the NDCT distribution via the trained model (Fig.\ref{fig:2}(c)).

\begin{figure}[htbp]
    \centering
    \includegraphics[width=0.5\textwidth]{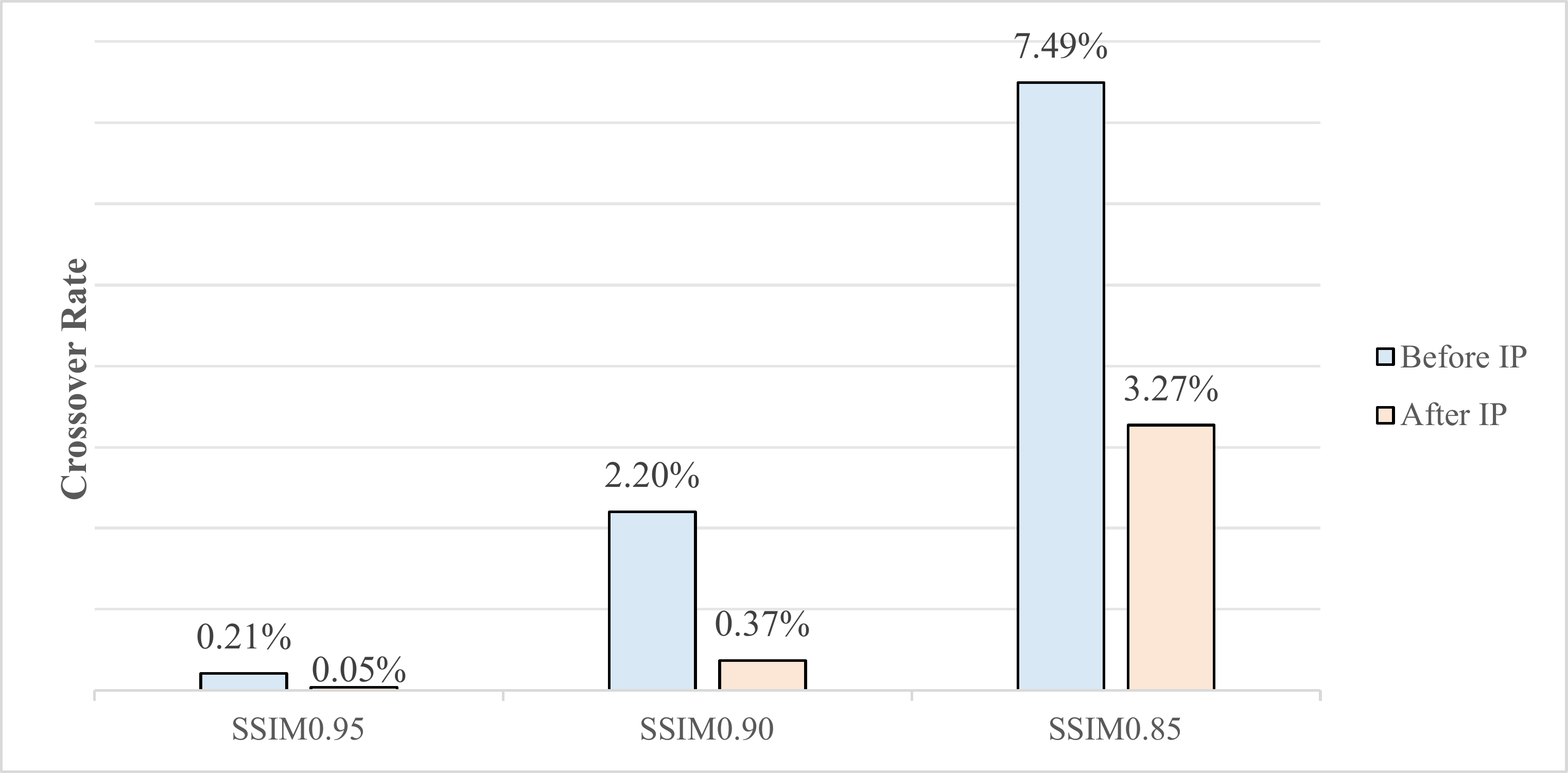}
    \captionsetup{width=\linewidth} 
    \caption{Effect of the Image Purification (IP) Strategy on Straightening Network Paths. We define the crossover rate as the proportion of samples with intersecting paths in the dataset. SSIM0.95, SSIM0.90, SSIM0.85 represent the cases where the similarity at the crossover point is $\geq$ 0.95, 0.90, 0.85, respectively, among the samples with crossovers. Details are in the \nameref{sec:补充材料}.}
    \label{fig:4}
\end{figure}

To quantify the "path straightening" effect of the IP strategy, we define a "crossing rate" to measure the severity of path crossings in the dataset. As shown in Fig.\ref{fig:4}, applying the IP strategy significantly reduces the crossing rate by a factor of 2 to 6 across different similarity thresholds, theoretically demonstrating that the IP strategy provides a superior data foundation for model training.

\begin{figure*}[b]
    \centering
    \includegraphics[width=\textwidth]{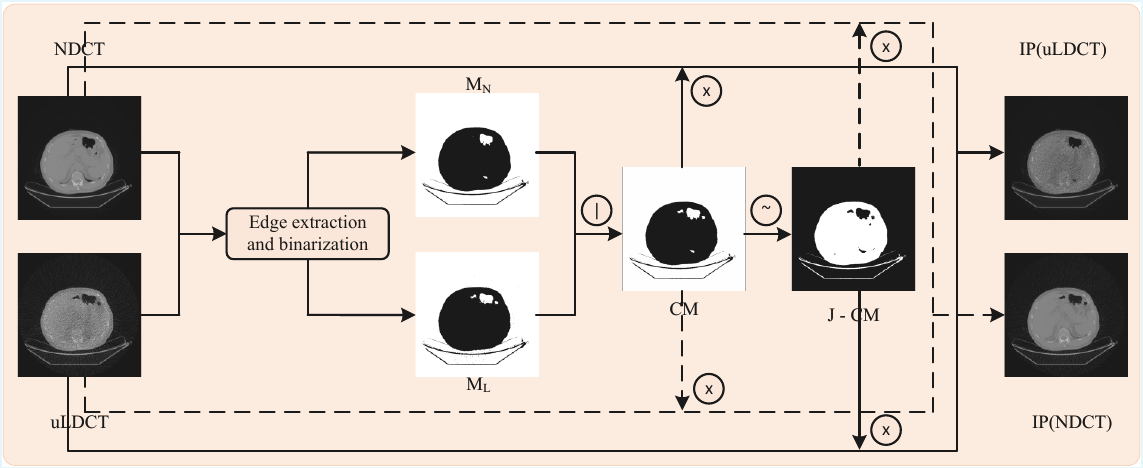}
    \captionsetup{width=\linewidth} 
    \caption{Proposed Image Purification (IP) strategy workflow.}
    \label{fig:5}
\end{figure*}

Our Image Purification (IP) strategy consists of five key steps, as illustrated in Fig.\ref{fig:5}: Edge Extraction \& Binarization, Common Mask Calculation, Image Purification, Training Data Selection, and Evaluation Label Selection.

\textbf{1.Edge Extraction \& Binarization.}This step applies Otsu's Method \cite{otsu1975threshold} to uLDCT and NDCT images for adaptive binarization, computing binary masks to separate the anatomical contours from the texture.

\textbf{2.Common Mask Calculation.}We use the binary masks $M_{L}$ (uLDCT) and $M_{N}$ (NDCT) from the previous step to calculate the Common Mask (CM) using Eq.\ref{eq:1}. Let $x$ denote any pixel location in the image. If the binary value in either mask is 1 (indicating the presence of an anatomical contour at that location), the value in the Common Mask at $x$ is set to 1; otherwise, it is 0.

\begin{equation}
  \operatorname{CM}(x)=\left(\operatorname{M}_{L}(x) \,|\, \operatorname{M}_{N}(x)\right)
  \label{eq:1}
\end{equation}

\textbf{3.Image Purification.}In Fig.\ref{fig:2}(b), we first decompose the residual vector $v$ from NDCT to uLDCT into two orthogonal components: one representing structural change ($v_1$) and one representing textural change ($v_2$). Specifically, we define:
\begin{equation}
  \mathrm{v}_1=\mathrm{X}_1{ }^{\prime}-\mathrm{X}_0 
  \label{eq:2}
\end{equation}
\begin{equation}
  \mathrm{v}_2=\mathrm{X}_1 - \mathrm{X}_1{ }^{\prime}
  \label{eq:3}
\end{equation}

where $X_0$ and $X_1$represent the NDCT and uLDCT images, respectively, and ${{X}_1 }^{\prime}$ is the intermediate image we wish to construct, structurally aligned with $X_0$.Using the Common Mask CM as the weight for anatomical contours and $J - CM$ (where $J$ is an all-ones matrix)as the weight for texture, we decompose the residual $v$ based on these two bases:

\begin{equation}
  \mathrm{v}_1= \left ( J - CM \right ) \cdot \mathrm{v}
  \label{eq:4}
\end{equation}
\begin{equation}
  \mathrm{v}_2= CM \cdot \mathrm{v}
  \label{eq:5}
\end{equation}

The resulting $v_1$ induces texture change while preserving structure, and $v_2$ induces structural change while preserving texture.Based on this decomposition, we generate the purified images using the following formulas:

\begin{equation}
  \mathrm{IP(uLDCT)}= \left ( J - CM \right ) \cdot \mathrm{uLDCT} \;+\; CM \cdot \mathrm{NDCT}
  \label{eq:6}
\end{equation}
\begin{equation}
  \mathrm{IP(NDCT)}= \left ( J - CM \right ) \cdot \mathrm{NDCT} \;+\; CM \cdot \mathrm{uLDCT}
  \label{eq:7}
\end{equation}

The IP(uLDCT) obtained via Eq.\ref{eq:6} possesses the anatomical contours of NDCT while retaining the texture of uLDCT. Similarly, IP(NDCT) obtained via Eq.\ref{eq:7} possesses the anatomical contours of uLDCT and the texture of NDCT.

To more finely control the distance from the constructed IP(uLDCT) distribution to the NDCT distribution, we further introduce a parameter $T$, proposing the following formulas:
\begin{align}
  \nonumber \mathrm{IP(uLDCT)}=& \left ( J - CM \right ) \cdot \left [\, \left ( 1-T \right )\cdot \mathrm{uLDCT} + T \cdot NDCT \, \right ]  \\& + CM \cdot \mathrm{NDCT}
  \label{eq:8}
\end{align}

Parameter $T$ controls the proportion of NDCT texture component in IP(uLDCT). When $T=0$, Eq.\ref{eq:8} is equivalent to Eq.\ref{eq:6}, and IP(uLDCT) is farthest from the NDCT distribution. When $T=1$, $IP(uLDCT)=NDCT$ and the two distributions completely overlap.

\textbf{4.Training Data Selection. }After IP preprocessing, we generate four core datasets: the original uLDCT and NDCT, and the purified IP(uLDCT) and IP(NDCT). Clear anatomical correspondences exist among these data: uLDCT shares anatomical structure with IP(NDCT), and NDCT shares anatomical structure with IP(uLDCT). Based on this, we construct three potential training data combinations to explore different learning paradigms:

Combination I: Use NDCT as the label and IP(uLDCT) as the input.

Combination II: Use IP(NDCT) as the label and uLDCT as the input.

Combination III: Mix the data from Combination I and Combination II.

To systematically evaluate the effectiveness of each combination, we conducted exhaustive comparative analyses in our ablation studies. Unless otherwise specified, all subsequent experiments in this paper default to using Combination I for model training.

\textbf{5.Evaluation Label Selection. }Since uLDCT and NDCT have different anatomical contours, using NDCT directly as the label for evaluating uLDCT denoising is unreasonable. We desire the denoised image to preserve the anatomical contours of the input uLDCT while recovering more texture details. The IP(NDCT) obtained via Eq.\ref{eq:7} possesses precisely the anatomical contours of the uLDCT and the texture of the NDCT, perfectly meeting this requirement. Therefore, in our framework, we use IP(NDCT) as the label for evaluating the denoised uLDCT images.
\subsection{FFM Model}
\label{sub:FFM模型}
Building upon the high-quality data provided by the IP strategy, we further propose a Frequency-domain Flow Matching (FFM) model to achieve superior structure-preserving denoising. We take Flow Matching \cite{FM} as the baseline and migrate its main network from the image domain to the frequency domain to further enhance its anatomical contour preservation capability.

\begin{figure}[ht]
    \centering
    \includegraphics[width=0.5\textwidth]{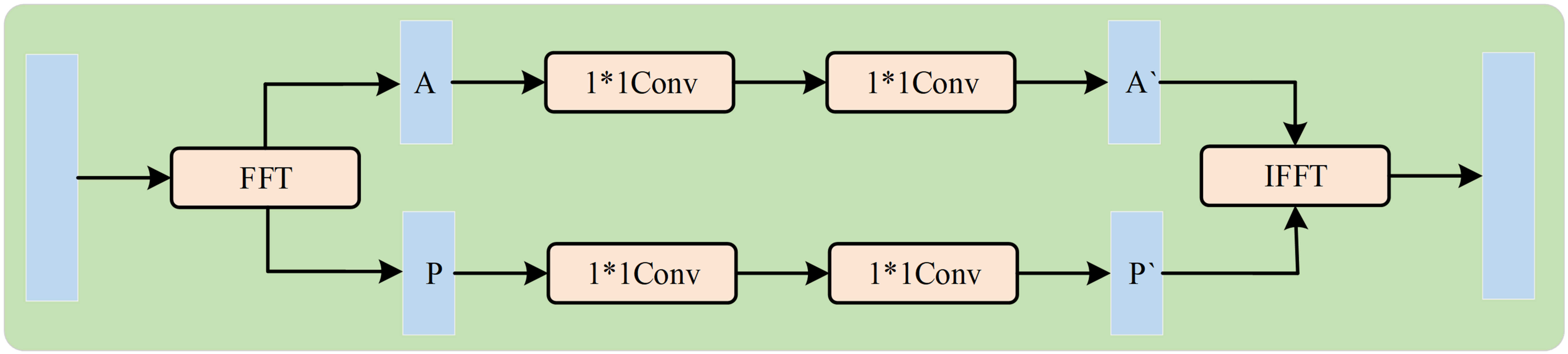}
    \captionsetup{width=\linewidth} 
    \caption{Frequency-domain module of FFM.}
    \label{fig:6}
\end{figure}

As shown in Fig.\ref{fig:6}, the frequency-domain module of FFM processes the input features at each layer of the U-Net backbone network. We employ a 2D Fast Fourier Transform (FFT) to transform the spatial information into the frequency domain, obtaining magnitude and phase components. Subsequently, we apply two separate 1x1 convolutions to these components to preserve the original structural information, followed by an Inverse Fast Fourier Transform (IFFT) to transform the features back to the spatial domain.

The training process of FFM follows the standard Flow Matching \cite{FM} paradigm, focusing on learning a velocity field from the noisy distribution to the clean image distribution (Fig.\ref{fig:2}). As shown in Fig.\ref{fig:3}, the proposed diffusion model includes a forward process and a reverse process. In the forward process, the model learns to predict the velocity field at any time $t$ based on the image $x_t$ and the time step $t$. The specific training procedure is outlined in Algorithm 1: In each iteration, we sample a pair of images $(x_0, x_1)$ from the paired NDCT and IP(uLDCT) set and sample a time step $t$ from the uniform distribution $U(0,1)$. Then, the intermediate state $x_t$ is computed using the linear interpolation formula Eq.\ref{eq:9}. The model's objective is to predict the velocity field $v_t = x_1 - x_0$ from $x_0 $ to $ x_1 $. We optimize the model parameters $\theta$ by minimizing the mean squared error loss function ${L}(\theta)$ between the predicted velocity field $\hat{v}_t$ and the true velocity field $v_t$.

During the inference stage, i.e., the reverse process, we use the trained model $R_{\theta}$ and a numerical method (e.g., Euler method) starting from the input uLDCT image $x_1 $ to iteratively solve until the final denoised image $x_0$ is obtained.

The linear interpolation formula for FFM in Fig.\ref{fig:3}, which corresponds to one-step noise addition in the forward process, is
\begin{equation}
    \mathrm{X}_{t}=D\left ( X_0,X_1,t \right )  =t \cdot X_1 \;+\; \left ( 1-t \right ) \cdot X_0
    \label{eq:9}
\end{equation}

The pseudo-code for the FFM training and sampling processes is presented in Algorithm \ref{alg:training} and Algorithm \ref{alg:sampling}, respectively.

\begin{algorithm}[H]
\caption{Training for FFM}
\label{alg:training}
\begin{algorithmic}[1]
\item[] \textbf{Input:} \hangindent=2em Paired NDCT and IP(uLDCT) image sets,$I = \{(x_0, x_1)_i\}_{i=1}^N$, time range $[0, 1]$ 
\item[] \textbf{Output:} Trained $R_{\theta}$
\STATE Initialization: Randomly initialize $R_{\theta}$
\REPEAT
\STATE Sample $(x_0, x_1) \sim I$
\STATE Sample $t \sim \text{Uniform}(0, 1)$
\STATE Calculate $x_t$ by linear interpolation: $x_t = t \cdot x_1 + (1-t) \cdot x_0$
\STATE Calculate $v_t$ by: $v_t = x_1 - x_0$
\STATE Calculate $\hat{v}_t$ by: $\hat{v}_t = R_{\theta}(x_t, t)$
\STATE Update $\theta$ by minimize $\mathcal{L}(\theta) = \mathbb{E}_{t,x_0,x_1} \|\hat{v}_t - v_t\|^2_2$
\UNTIL{converged}
\end{algorithmic}
\end{algorithm}
\begin{algorithm}[H]
\caption{Sampling for FFM (Euler Method)}
\label{alg:sampling}
\begin{algorithmic}[1]
\item[] \textbf{Input:} \hangindent=2em A test uLDCT image $x_1$, number of steps $T$, trained model $R_{\theta}$
\item[] \textbf{Output:} Denoised image $x_0$
\STATE Load the trained $R_{\theta}$
\STATE Set $x = x_1$
\STATE Set $\Delta t = 1/T$
\FOR{$t = 1, 1-\Delta t, \ldots, 0$}
\STATE Calculate $\hat{v}$ by: $\hat{v} = R_{\theta}(x, t)$
\STATE Update: $x = x - \Delta t \cdot \hat{v}$
\ENDFOR
\STATE $x_0 = x$
\STATE \textbf{return} $x_0$
\end{algorithmic}
\end{algorithm}

\section{EXPERIMENTS}
\label{sec:EXPERIMENTS}

\subsection{Our Real-World Datasets}
\label{sub:Our Real-World Datasets}
To validate the effectiveness of the proposed method, we conducted comprehensive experimental evaluations on a carefully collected real-world clinical dataset. This dataset originates from 6 patients, containing 4310 pairs of uLDCT and NDCT images. We randomly split the data approximately 7:1.5:1.5, resulting in 3017 pairs for training, 646 for validation, and 647 for testing.

Regarding scanning parameters, the normal-dose CT used a tube voltage of 120 kV and current of 250 mA, while the uLDCT used significantly reduced settings of 80 kV and 10 mA. This drastic dose reduction resulted in severe noise and artifacts in the images. Calculations confirm that the radiation dose for uLDCT scans is approximately 2\% of that for NDCT scans, firmly placing it in the ultra-low-dose category and presenting unprecedented challenges for denoising algorithms.

During the data preprocessing stage, we applied the aforementioned Image Purification strategy to each uLDCT and NDCT image pair in the training set, generating structurally aligned IP(uLDCT) and IP(NDCT) image pairs. This image-level purification ensures no loss in dataset size, thereby supporting networks trained on both full images and image patches.
\subsection{Experimental Settings}
\label{sub:实验设置}
\textbf{Implementation Details.}
To ensure fairness and reproducibility, we standardized the training details for all compared models. For networks whose authors publicly provided training code, we used their default settings directly. For networks without available training code, we uniformly used the Adam optimizer \cite{Adam} with an initial learning rate of $10^{-4}$ and a batch size of 2. All experiments were conducted on an NVIDIA RTX 4090 GPU. For our proposed FFM model, the training was set for 200 epochs, and the Euler method with 10 time steps was used for inference. For other diffusion models, we used the sampling steps provided by the original authors: IPDM \cite{IPDM} and Cold Diffusion \cite{coldDiffusion} used 50 steps; CoreDiff \cite{corediff} and Flow Matching \cite{FM} used 10 steps, consistent with FFM; for Resshift \cite{resshift}, we experimented with 15 and 4 steps and ultimately selected 15 steps to ensure sampling quality. Other non-diffusion models can be considered single-step denoisers (time step = 1).

\textbf{Evaluation Metrics.}
To comprehensively and objectively evaluate denoising performance, we employed seven widely recognized image quality metrics. In addition to traditional SSIM, PSNR, and RMSE, we introduced FSIM, GMSD, VIF, and NQM. Previous research suggests that FSIM, VIF, and NQM have higher consistency with radiologists' subjective evaluations.\cite{radiologists’favor} Therefore, we divided these seven metrics into two groups: The first group includes FSIM, GMSD, and SSIM, primarily used to assess the model's ability to recover anatomical contours. The second group includes VIF, NQM, PSNR, and RMSE, focusing on evaluating the model's ability to recover texture details. In experiments validating the effectiveness of the IP strategy, we primarily focused on the first group of metrics. In ablation studies investigating the parameter T, we considered all seven metrics comprehensively.
\subsection{Comparison on Real-World uLDCT Images}
\label{sub:真实世界LDCT图像对比}
To comprehensively evaluate the effectiveness of the proposed framework, we systematically compared various denoising networks trained with our IP strategy against other representative methods on our constructed real-world patient lung dataset.

\begin{table*}[htbp]
\centering
\setlength{\tabcolsep}{1pt} 
\newcommand{\optibold}[1]{\scalebox{0.92}{\textbf{#1}}} 
\captionsetup{width=\textwidth}
\caption{Quantitative Results (Mean $\pm$ Standard Deviation) of Different Algorithms on the Real Patient Lung Dataset.Improvement rates over the baseline method in the previous column are shown as percentages.\textcolor{red}{Red}, \textcolor{blue}{blue}, and \underline{underlined} text indicate the best, second-best, and third-best results, respectively.}
\label{tab:results}
\begin{tabularx}{\textwidth}{@{}l c >{\centering\arraybackslash}X c >{\centering\arraybackslash}X c >{\centering\arraybackslash}X@{}}
\toprule[0.5mm]
\textbf{Method} & \multicolumn{2}{c}{\textbf{FSIM$\uparrow$}} & \multicolumn{2}{c}{\textbf{GMSD$\downarrow$}} & \multicolumn{2}{c}{\textbf{SSIM$\uparrow$}} \\
\midrule[0.5mm]
uLDCT & & 0.5020$\pm$0.0717 & & 0.7196$\pm$0.0130 & & 0.4788$\pm$0.1197  \\

\rowcolor{gray!20}uLDCT + IP & 57\% & 0.7889$\pm$0.0871 & 18\% & 0.5870$\pm$0.0982 & 63\% & 0.7814$\pm$0.1118 \\ 

flow matching\cite{FM} & & 0.8556$\pm$0.0269 & & 0.5132$\pm$0.0307 & & 0.8740$\pm$0.0651 \\

\rowcolor{gray!20}flow matching\cite{FM} + IP & 8\% & \underline{0.9262$\pm$0.0359} & 28\% & \textcolor{red}{0.3691$\pm$0.0627} & 5\% & \textcolor{blue}{0.9160$\pm$0.0639} \\

CoreDiff\cite{corediff} & & 0.8309$\pm$0.0500 & & 0.6977$\pm$0.0340 & & \underline{0.9134$\pm$0.0408} \\

\rowcolor{gray!20}CoreDiff\cite{corediff} + IP & 8\% & 0.8958$\pm$0.0632 & 25\% & 0.5249$\pm$0.1046 & -1\% & 0.9057$\pm$0.0737 \\

Cold Diffusion\cite{coldDiffusion} & & 0.8706$\pm$0.0219 & & 0.5205$\pm$0.0482 & & 0.8970$\pm$0.0547 \\

\rowcolor{gray!20}Cold Diffusion\cite{coldDiffusion} + IP & 7\% & \textcolor{blue}{0.9308$\pm$0.0302} & 27\% & \underline{0.3788$\pm$0.0551} & 2\% & 0.9120$\pm$0.0577 \\

REDCNN\cite{redcnn} & & 0.8047$\pm$0.0339 & & 0.6941$\pm$0.0728 & & 0.8674$\pm$0.0472 \\

\rowcolor{gray!20}REDCNN\cite{redcnn} + IP & 12\% & 0.8985$\pm$0.0538 & 39\% & 0.4201$\pm$0.0896 & 3\% & 0.8936$\pm$0.0744 \\

EDCNN\cite{edcnn} & & 0.7561$\pm$0.0382 & & 0.6625$\pm$0.0464 & & 0.8192$\pm$0.0373 \\

\rowcolor{gray!20}EDCNN\cite{edcnn} + IP & 16\% & 0.8782$\pm$0.0515 & 30\% & 0.4615$\pm$0.0790 & 6\% & 0.8706$\pm$0.0723 \\

DuGAN\cite{DU-GAN} & & 0.7964$\pm$0.0361 & & 0.5210$\pm$0.0340 & & 0.8369$\pm$0.0559 \\

\rowcolor{gray!20}DuGAN\cite{DU-GAN} + IP & 14\% & 0.9107$\pm$0.0396 & 25\% & 0.3931$\pm$0.0627 & 6\% & 0.8900$\pm$0.0761 \\

MMCA\cite{MMCA} & & 0.6478$\pm$0.0325 & & 0.6026$\pm$0.0533 & & 0.8025$\pm$0.0607 \\

\rowcolor{gray!20}MMCA\cite{MMCA} + IP & 38\% & 0.8935$\pm$0.0505 & 28\% & 0.4317$\pm$0.0839 & 11\% & 0.8884$\pm$0.0781 \\
\hline
N2N\cite{N2N} & & 0.5406$\pm$0.0902 & & 0.7215$\pm$0.0278 & & 0.5484$\pm$0.1546 \\

IPDM\cite{IPDM} & & 0.5205$\pm$0.0656 & & 0.7346$\pm$0.0176 & & 0.5316$\pm$0.1264 \\

Resshift\cite{resshift} & & 0.8073$\pm$0.0206 & & 0.6864$\pm$0.0313 & & 0.7538$\pm$0.0290 \\

FFM(ours) + IP & & \textcolor{red}{0.9342$\pm$0.0341} & & \textcolor{blue}{0.3708$\pm$0.0665} & & \textcolor{red}{0.9230$\pm$0.0625} \\
\bottomrule[0.5mm]
\label{table:1}
\end{tabularx}
\end{table*}
\subsubsection{Comparison Methods}
\label{subsub:比较方法}
This comparative experiment aims to evaluate the effectiveness of the proposed framework from multiple dimensions, specifically:

\begin{enumerate}
    \item We first compared the structural similarity between the original uLDCT images and the uLDCT images processed by the IP strategy against the NDCT images, respectively, to assess the improvement in data quality brought by the IP strategy itself.
    \item  We selected various mainstream LDCT denoising networks, including CoreDiff \cite{corediff}, REDCNN \cite{redcnn}, EDCNN \cite{edcnn}, DuGAN \cite{DU-GAN}, and MMCA \cite{MMCA}, as well as general image denoising networks like Flow Matching \cite{FM} and Cold Diffusion \cite{coldDiffusion} . We evaluated their performance on our real-world patient lung dataset, both on the original data and when combined with our IP strategy. 
    \item We also evaluated self-supervised methods that do not require paired data for training, namely IPDM \cite{IPDM} and N2N \cite{N2N}, as well as a supervised general image denoising network, Resshift \cite{resshift} .
    \item Finally, we combined our proposed Frequency-domain Flow Matching model (FFM) with the IP strategy, trained and evaluated it on the same dataset, to verify its comprehensive performance.
\end{enumerate}

\subsubsection{Quantitative Results}
\label{subsub:客观结果}
The results in Table \ref{table:1} demonstrate the universal effectiveness of our IP strategy in enhancing the performance of various LDCT denoising models. Compared to their baselines, all methods showed consistent improvement in contour preservation metrics (FSIM, GMSD, SSIM) when combined with the IP strategy. Notably, simpler models like REDCNN \cite{redcnn} and EDCNN \cite{edcnn} exhibited greater improvement margins. This phenomenon indicates that for the uLDCT denoising task, addressing the quality of the training data (i.e., structural alignment) is a foundational prerequisite for model performance, whose importance may even surpass the complexity of the model architecture itself. Powerful generative models like Flow Matching \cite{FM} and Cold Diffusion \cite{coldDiffusion}, with the assistance of the IP strategy, can also further break through performance bottlenecks and achieve higher denoising quality.

The performance differences among models also reflect the challenges of uLDCT denoising: self-supervised methods IPDM \cite{IPDM} and N2N \cite{N2N} performed poorly due to the lack of explicit supervision signals; the latent diffusion model (Resshift \cite{resshift}), by operating in a low-dimensional latent space, inevitably loses some high-frequency detail information during the encoding-decoding process, which may limit its performance in uLDCT tasks requiring high-precision structural recovery. Our proposed FFM model, empowered by the IP strategy, achieved the best overall performance, validating the effectiveness of our framework.

To directly assess the impact of the IP strategy on data quality, we first compared the structural similarity between the original uLDCT images and the uLDCT images processed by the IP strategy (uLDCT + IP) against the NDCT images. The results are detailed in Table \ref{table:1}. This directly proves the effectiveness of the IP strategy in improving data quality, laying a solid foundation for the subsequent significant performance improvement of the models.

Building on this, the quantitative results show that our IP strategy has significant universality in enhancing the performance of various LDCT denoising models. Compared to the baseline models without the IP strategy, all methods achieved consistent improvements in contour preservation metrics (FSIM, GMSD, SSIM). A particularly noteworthy phenomenon is that structurally simpler models like REDCNN \cite{redcnn} and EDCNN \cite{edcnn}, when combined with the IP strategy, showed much greater performance improvements than powerful generative models like Flow Matching \cite{FM} and Cold Diffusion \cite{coldDiffusion}. This finding profoundly reveals that for the extreme task of uLDCT denoising, solving the quality issue of the training data (i.e., structural alignment) is a foundational prerequisite for model performance, whose importance may even exceed the complexity of the model architecture itself. Powerful generative models like Flow Matching \cite{FM} and Cold Diffusion \cite{coldDiffusion}, with the assistance of the IP strategy, can also further break through performance bottlenecks and achieve higher denoising quality.

The performance differences among models further reflect the inherent challenges of the uLDCT denoising task: self-supervised methods IPDM \cite{IPDM} and N2N \cite{N2N} yielded unsatisfactory denoising results due to the lack of clear paired supervision signals. The latent diffusion model Resshift \cite{resshift}, due to its operation in a low-dimensional latent space, inevitably loses part of the high-frequency detail information during the encoding-decoding process, limiting its performance in uLDCT tasks requiring high-precision structural recovery. Our proposed FFM model, empowered by the IP strategy, achieved the best comprehensive performance, validating the effectiveness of our framework.

\begin{figure*}[ht]
    \centering
    \includegraphics[width=\columnwidth]{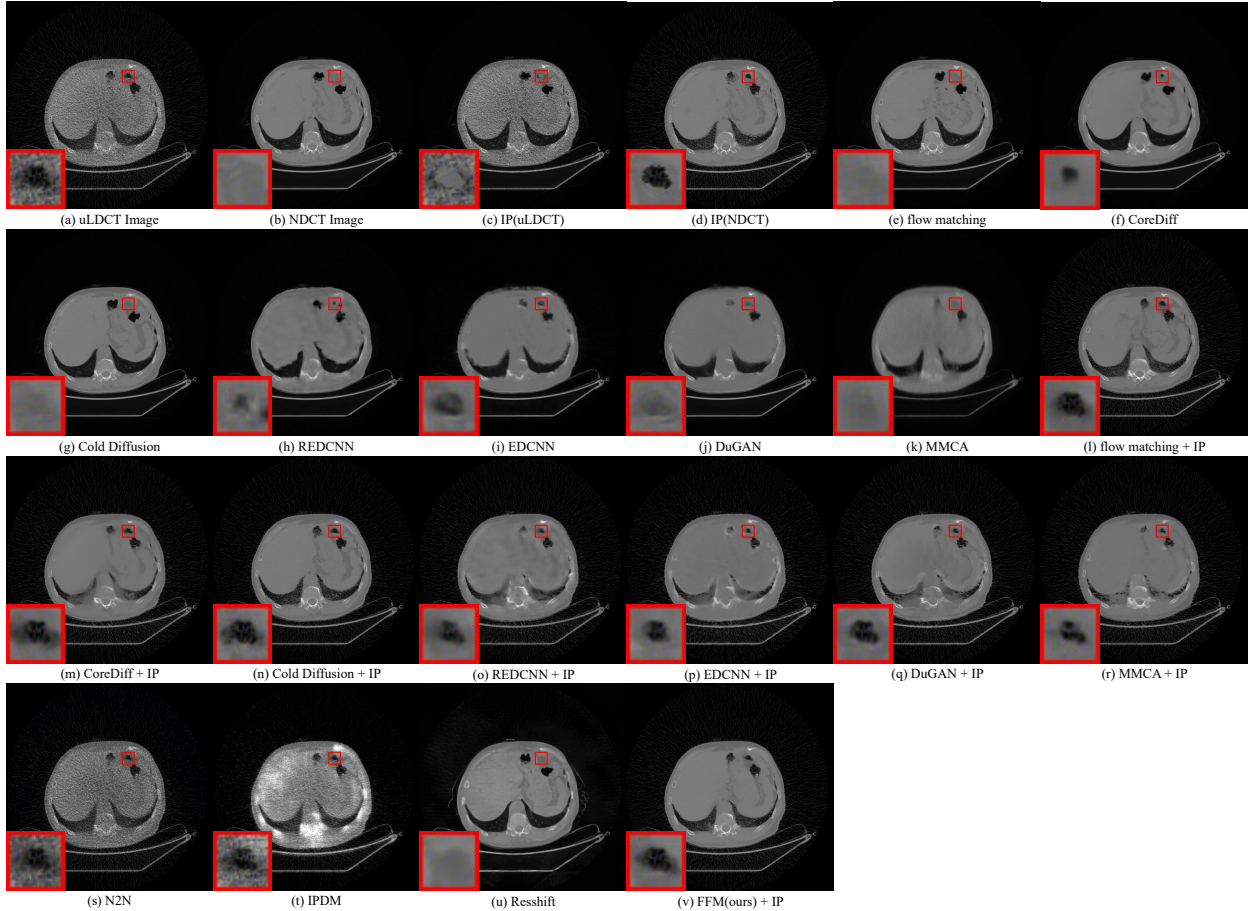}
    \captionsetup{width=\linewidth} 
    \caption{Comparison of visual quality by different denoising methods on one uLDCT image from our Patient dataset.}
    \label{fig:7}
\end{figure*}

\subsubsection{Visual Quality}
\label{subsub:视觉质量}
Visual comparison results provide intuitive and strong evidence for our quantitative analysis, clearly showing the performance differences among methods in preserving anatomical structures. First, our IP strategy itself effectively corrects structural misalignment between images. As shown in Fig.\ref{fig:7}(c) and Fig.\ref{fig:7}(d), for anatomical contours present in uLDCT but absent in NDCT, the IP strategy corrects the uLDCT by filling in that contour; conversely, for contours present in NDCT but absent in uLDCT, the IP strategy corrects the NDCT by creating that contour, ultimately making the corrected image pair have completely consistent anatomical structures.

In models trained without the IP strategy, the denoising results generally exhibited structural distortion. As shown in Fig.\ref{fig:7}(e)-(k), specifically, Flow Matching \cite{FM}, Cold Diffusion \cite{coldDiffusion}, and MMCA \cite{MMCA} tended to fill in the anatomical contours present in the uLDCT; CoreDiff \cite{corediff} and REDCNN \cite{redcnn} caused deformation of some anatomical contours; while EDCNN \cite{edcnn} and DuGAN \cite{DU-GAN} blurred the original anatomical contours. These methods altered the original tissue structure of the input uLDCT image to varying degrees.

In contrast, all models trained with data pairs corrected by the IP strategy successfully preserved the anatomical contours in the uLDCT, as shown in Fig. \ref{fig:7}(l)-(r). This fully demonstrates the key role of the IP strategy in ensuring structural consistency in the models. Among the self-supervised methods, N2N \cite{N2N} showed almost no denoising capability, while IPDM \cite{IPDM} produced over-exposed results. These outcomes highlight the inherent limitations of self-supervised methods when dealing with the extreme noise and complex artifacts in uLDCT. The output images of Resshift \cite{resshift} contained obvious horizontal stripe artifacts. This phenomenon reveals a key bottleneck of methods based on latent diffusion models in the uLDCT denoising field: the encoding-decoding process inevitably leads to the loss of high-frequency detail information.

Furthermore, compared to other methods combined with the IP strategy in Fig.\ref{fig:7}(l)-(n), our FFM model shown in Fig.\ref{fig:7}(v) preserved more complete and clearer anatomical contour information from the uLDCT. This validates the synergistic advantage of combining frequency-domain processing with high-quality data in achieving superior structural fidelity. More visual quality comparison results are provided in the \nameref{sec:补充材料}.
\subsection{Ablation Studies}
\label{sub:消融研究}
To systematically verify the effectiveness of key components in our framework, we conducted a series of thorough ablation studies, focusing primarily on the following three core issues:

\textbf{1.Effect of Parameter T:} We investigated the impact of parameter T in the IP strategy on the performance of uLDCT image denoising networks. Parameter T controls the proportion of NDCT texture component in the constructed IP(uLDCT) image. We trained the Flow Matching model \cite{FM} on the patient dataset while varying the value of T. The experimental results are shown in Table \ref{table:2}. The results indicate that when T=0.0, the model achieved the optimal SSIM value; when T=0.1, the model performed best on five metrics: FSIM, VIF, NQM, PSNR, and RMSE. This finding suggests that moderately introducing NDCT texture information into the constructed uLDCT image helps improve the comprehensive performance of the denoising network.

\begin{table}[H]
\centering
\caption{Results of Training FM\cite{FM} on Patient Data with Different T Values.}
\label{tab:different_t}
\begin{adjustbox}{width=0.8\columnwidth}
\begin{tabular}{cccccccc}
\toprule[0.5mm]
\textbf{T} & \textbf{FSIM$\uparrow$} & \textbf{GMSD$\downarrow$} & \textbf{SSIM$\uparrow$} & \textbf{VIF$\uparrow$} & \textbf{NQM$\uparrow$} & \textbf{PSNR$\uparrow$} & \textbf{RMSE$\downarrow$} \\
\midrule[0.5mm]
0.0   & \textcolor{blue}{0.9294}   & \textcolor{blue}{0.3663}   & \textcolor{red}{0.9183}   & \textcolor{blue}{0.8063}   & \textcolor{blue}{19.83}   & \textcolor{blue}{30.88}   & \textcolor{blue}{8.0}   \\
0.1   & \textcolor{red}{0.9305}   & 0.3735   & \textcolor{blue}{0.9170}   & \textcolor{red}{0.8603}   & \textcolor{red}{20.76}   & \textcolor{red}{31.49}   & \textcolor{red}{7.4}   \\
0.2   & 0.9269   & 0.3924   & 0.9099   & 0.7337   & 17.51   & 29.15   & 9.8   \\
0.3   & 0.9277   & \textcolor{red}{0.3539}   & 0.9062   & 0.7659   & 17.67   & 29.19   & 9.7   \\
0.4   & 0.9252   & 0.3689   & 0.8936   & 0.7981   & 17.87   & 29.28   & 9.6   \\
0.5   & 0.9180   & 0.3917   & 0.8827   & 0.7558   & 16.36   & 28.06   & 11.0  \\
\bottomrule[0.5mm]
\label{table:2}
\end{tabular}
\end{adjustbox}
\end{table}

\begin{figure}[htbp]
    \centering
    \includegraphics[width=\columnwidth]{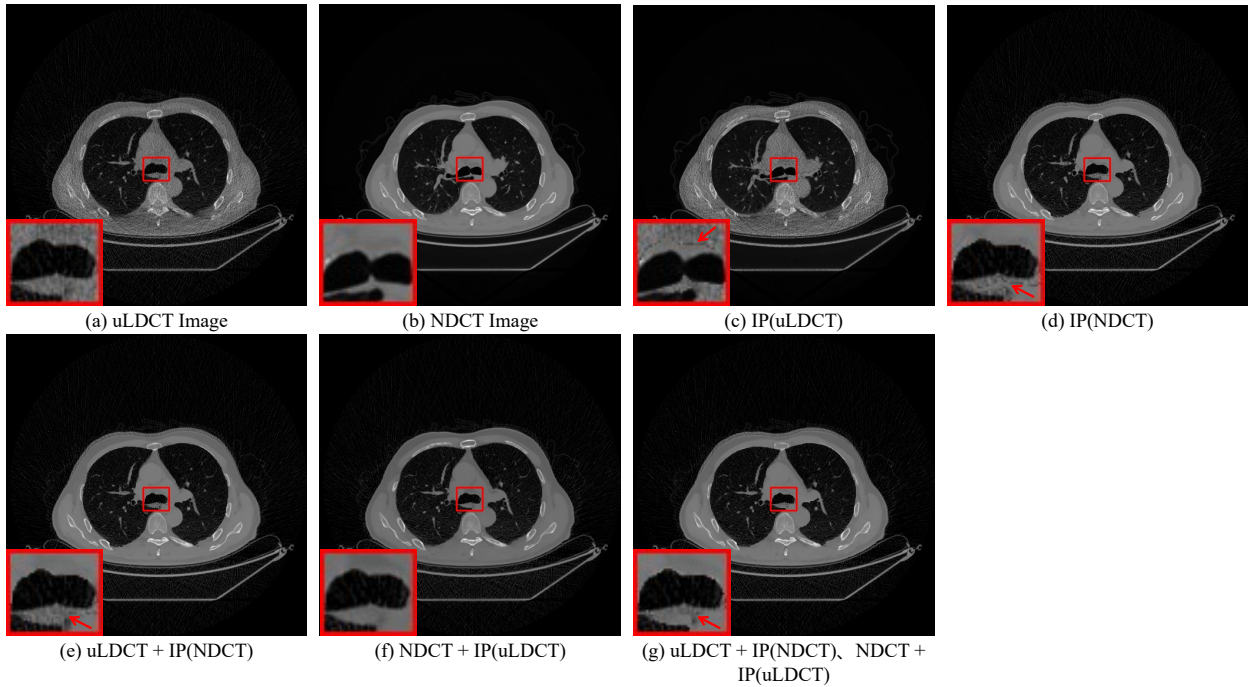}
    \captionsetup{width=\linewidth} 
    \caption{Comparison of Visual Quality by Different Training Data Pairs on One uLDCT Image from Our Patient Dataset. (a) to (d) are training data.(e) to (g) are sampling results using flow matching\cite{FM} under different training data, respectively.The \textcolor{red}{red} arrow indicates the overlapping trace of contours after using the IP strategy on the two images.}
    \label{fig:8}
\end{figure}

\textbf{2.Selection of Training Data Pairs:}We evaluated the impact of different training data pairs on model performance. The application of the IP strategy yields three potential training data combinations: (1) uLDCT with IP(NDCT); (2) NDCT with IP(uLDCT); (3) A mixture of the first two. As shown in Fig.\ref{fig:8}(c) and (d), the IP strategy may introduce slight granularity at the overlapping edges of the anatomical contours in the two images. The experimental results clearly show that the model trained using Combination (2) (Fig.\ref{fig:8}(f)) has superior visual quality compared to those using Combination (1) (Fig.\ref{fig:8}(e)) or Combination (3) (Fig.\ref{fig:8}(g)), both of which exhibit similar edge granularity. The fundamental reason is that Combinations (1) and (3) establish a mapping from uLDCT to IP(NDCT), causing the model to sample from the IP(NDCT) distribution during inference, thereby deviating from the target NDCT distribution.

\textbf{3.Choice of Processing Domain:}Finally, we verified the effectiveness of performing uLDCT image denoising in the frequency domain. In the image domain, signal and noise are highly coupled, making effective separation difficult and limiting the performance ceiling of denoising networks. To address this, we proposed the FFM model and compared it with the image-domain baseline model Flow Matching \cite{FM}. As shown in Table \ref{table:3}, FFM significantly outperformed FM on all six other metrics except GMSD. This result strongly proves that migrating the denoising process to the frequency domain can effectively enhance the model's comprehensive performance.

\begin{table}[H]
\centering
\caption{Results of Training FM\cite{FM} in Different Domains on the Patient Dataset.}
\label{tab:different_domains}
\begin{adjustbox}{width=0.8\columnwidth}
\begin{tabular}{lccccccc}
\toprule[0.5mm]
\textbf{Domain} & \textbf{FSIM$\uparrow$} & \textbf{GMSD$\downarrow$} & \textbf{SSIM$\uparrow$} & \textbf{VIF$\uparrow$} & \textbf{NQM$\uparrow$} & \textbf{PSNR$\uparrow$} & \textbf{RMSE$\downarrow$} \\
\midrule[0.5mm]
Frequency & \textcolor{red}{0.9342} & 0.3708 & \textcolor{red}{0.9230} & \textcolor{red}{0.9054} & \textcolor{red}{21.78} & \textcolor{red}{32.25} & \textcolor{red}{6.9} \\
Image     & 0.9292 & \textcolor{red}{0.3587} & 0.9179 & 0.8964 & 21.37 & 31.85 & 7.1 \\
\bottomrule[0.5mm]
\label{table:3}
\end{tabular}
\end{adjustbox}
\end{table}

\section{CONCLUSION}
\label{sec:CONCLUSION}

This paper addressed the problem of anatomical structure distortion in real-world ultra-low-dose CT (uLDCT) denoising caused by spatial data misalignment by proposing a systematic solution based on an Image Purification strategy. We constructed a real clinical dataset with a radiation dose of only 2\% of the normal level and introduced an innovative Image Purification strategy that effectively resolves the inherent spatial misalignment in real-world data. Building upon this, our proposed Frequency-domain Flow Matching model, working synergistically with the IP strategy, achieved state-of-the-art performance in preserving anatomical structures. This research provides new ideas and a powerful tool for promoting the clinical translation of uLDCT denoising technology.
\clearpage

\section*{Supplementary Material}
\label{sec:补充材料}
\appendix
In this Supplemental Material we define and compute the crossing rate (Sec.\ref{sec:Calculation of the Crossing Rate}), compare PSP\cite{psp} and our IP strategy quantitatively (Sec.\ref{sec:Quantitative Comparison with the PSP Strategy}), and show extra qualitative/visual results (Sec.\ref{sec:Additional Qualitative Results}).

\section{Calculation of the Crossing Rate}
\label{sec:Calculation of the Crossing Rate}

The crossing rate is a novel metric introduced in this paper to quantitatively assess the degree of intersection between paired data‑mapping trajectories. While the main text offers a concise definition, the following exposition provides a more comprehensive treatment. Let $x_1,x_2$ denote samples drawn from the NDCT distribution and let $y_1,y_2$ denote samples drawn from the uLDCT distribution, with the pairs $(x_1,y_2)$ and $(x_2,y_2)$ established accordingly. In a multi‑step sampling model, if the similarity between $y_1$ and $y_2$ progressively increases as the sampling proceeds, the model tends to generate predictions that gravitate toward the mean of the two samples. Consequently, the discrepancy between the predicted values $\hat{x}_1, \hat{x}_2$ and the true labels $x_1,x_2$ is amplified. By contrast, in a single‑step sampling model the intersection of trajectories occurs implicitly, yet it similarly enlarges the gap between predictions and ground‑truth labels.

Under the idealized setting of a multi‑step sampling model, the prediction at time $t$ is expressed as
\begin{equation}
  \mathrm{z}_t = \mathrm{t} \, \mathrm{x} + (1 - \mathrm{t}) \, \mathrm{y}
  \label{eq:11}
\end{equation}

where $z_t$ represents the model’s output at the intermediate stage $t$. If the condition
\begin{equation}
  \mathrm{SSIM}(y_1, y_2) < \mathrm{SSIM}(z_{t_1}, z_{t_2})
  \label{eq:12}
\end{equation}
holds, the samples associated with $y_1$ and $y_2$ are deemed to have intersecting mapping paths. The proportion of samples satisfying this inequality relative to the entire dataset defines the \textbf{crossing rate}.

Moreover, when the above inequality is satisfied simultaneously with
\begin{equation}
  \mathrm{SSIM}(x_1, x_2) < \mathrm{SSIM}(z_{t_1}, z_{t_2})
  \label{eq:13}
\end{equation}
the trajectories $z_{t_1}, z_{t_2}$ exhibit a more pronounced misleading effect on the model. All crossing‑rate values reported in this work are computed based on samples that meet both criteria.

It is important to note that these conclusions presuppose
\begin{equation}
  \mathrm{SSIM}(z_{t_1}, z_{t_2})  \geq  p
  \label{eq:14}
\end{equation}
where the threshold $p$ is chosen close to 1 to ensure that $z_{t_1}, z_{t_2}$ possess the capacity to “deceive” the model. In the figures of the main text, three values of $p$ are examined: 0.95, 0.90, and 0.85.

\section{Quantitative Comparison with the PSP Strategy}
\label{sec:Quantitative Comparison with the PSP Strategy}

The PSP\cite{psp} strategy is a conventional data‑cleaning method designed for low‑dose CT denoising. However, in the ultra‑low‑dose CT scenario, PSP\cite{psp} tends to discard a substantial amount of fine‑grained texture information. To address this limitation, we propose the IP strategy, which is specifically crafted for ultra‑low‑dose CT denoising. IP aims to preserve structural consistency while retaining subtle textures.

\begin{table*}[htbp]
\centering
\setlength{\tabcolsep}{1pt} 
\newcommand{\optibold}[1]{\scalebox{0.92}{\textbf{#1}}} 
\captionsetup{width=\textwidth}
\caption{Quantitative comparison of baseline, integrated IP strategy and PSP\cite{psp} strategy}
\label{tab:results}
\begin{tabularx}{\textwidth}{@{}l c >{\centering\arraybackslash}X c >{\centering\arraybackslash}X c >{\centering\arraybackslash}X@{}}
\toprule[0.5mm]
\textbf{Method} & \multicolumn{2}{c}{\textbf{FSIM$\uparrow$}} & \multicolumn{2}{c}{\textbf{GMSD$\downarrow$}} & \multicolumn{2}{c}{\textbf{SSIM$\uparrow$}} \\
\midrule[0.5mm]

flow matching\cite{FM} & & 0.8556$\pm$0.0269 & & 0.5132$\pm$0.0307 & & 0.8740$\pm$0.0651 \\

\rowcolor{gray!20}flow matching\cite{FM} + IP & 8\% & 0.9262$\pm$0.0359 & 28\% & 0.3691$\pm$0.0627 & 5\% & 0.9160$\pm$0.0639 \\

\rowcolor{red!20}flow matching\cite{FM} + PSP\cite{psp} & -25\% & 0.6433$\pm$0.0374 & -26\% & 0.6445$\pm$0.0363 & -28\% & 0.6282$\pm$0.0381 \\

DuGAN\cite{DU-GAN} & & 0.7964$\pm$0.0361 & & 0.5210$\pm$0.0340 & & 0.8369$\pm$0.0559 \\

\rowcolor{gray!20}DuGAN\cite{DU-GAN} + IP & 14\% & 0.9107$\pm$0.0396 & 25\% & 0.3931$\pm$0.0627 & 6\% & 0.8900$\pm$0.0761 \\

\rowcolor{red!20}DuGAN\cite{DU-GAN} + PSP\cite{psp} & -1\% & 0.7915$\pm$0.0388 & -1\% & 0.5257$\pm$0.0302 & 0.3\% & 0.8397$\pm$0.0577 \\

\bottomrule[0.5mm]
\label{table:4}
\end{tabularx}
\end{table*}

Table \ref{table:4} compares the denoising performance of two model families under different strategies. For the flow‑matching model (a multi‑step sampler), applying the PSP\cite{psp} strategy reduces denoising capability by more than 25\% relative to the baseline that employs no strategy. In contrast, the DuGAN \cite{DU-GAN} model (a single‑step sampler) experiences only a marginal decline when PSP\cite{psp} is used. Qualitative analysis reveals that single‑step denoising models such as DuGAN \cite{DU-GAN} inherently possess limited ability to recover texture details; consequently, the adverse impact of PSP\cite{psp} on their performance is negligible.


\section{Additional Qualitative Results}
\label{sec:Additional Qualitative Results}
Because of space constraints in the main manuscript, this section presents a supplementary set of qualitative comparisons. The conclusions drawn from these additional results are fully consistent with those reported in the main text, thereby providing further evidence for the effectiveness of the proposed IP strategy and the FFM model.

\begin{figure*}[htbp]
    \centering
    \includegraphics[width=0.8\columnwidth]{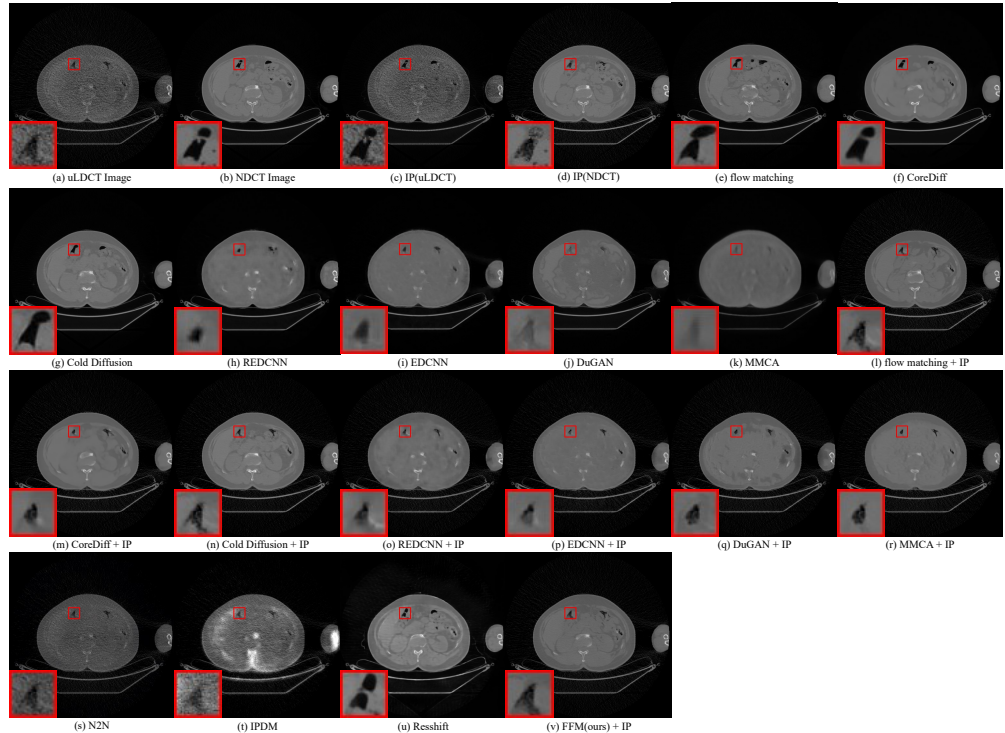}
    \captionsetup{width=0.8\linewidth} 
    \caption{Comparison of Visual Quality by Different Denoising Methods on One uLDCT Image from Our Patient Dataset.}
    \label{fig:11}
\end{figure*}

\begin{figure*}[htbp]
    \centering
    \includegraphics[width=0.8\columnwidth]{figs/12.pdf}
    \captionsetup{width=0.8\linewidth} 
    \caption{Comparison of Visual Quality by Different Denoising Methods on One uLDCT Image from Our Patient Dataset.}
    \label{fig:12}
\end{figure*}

\begin{figure*}[htbp]
    \centering
    \includegraphics[width=0.8\columnwidth]{figs/13.pdf}
    \captionsetup{width=0.8\linewidth} 
    \caption{Comparison of Visual Quality by Different Denoising Methods on One uLDCT Image from Our Patient Dataset.}
    \label{fig:13}
\end{figure*}

\clearpage

\end{document}